\documentclass{article}
\usepackage[final,dblblindworkshop,nonatbib]{neurips_2026}
\usepackage[T1]{fontenc}
\usepackage{booktabs}
\usepackage{array,multirow}
\usepackage{amsmath,amssymb}
\usepackage{xcolor}
\usepackage{graphicx}
\usepackage{enumitem}
\usepackage{float}
\usepackage[hidelinks]{hyperref}

\newcommand{\code}[1]{\texttt{#1}}
\renewcommand{\to}{\ensuremath{\rightarrow}}
\newcommand{\releaseurl}{https://github.com/Evaaaz/RTLRepairBench}

\title{What Do RTL Repair Benchmarks Actually Measure?}
\workshoptitle{AI for Chip Design}
\author{%
  Yeyin (Eva) Zhu\thanks{Equal contribution.}\\
  NVIDIA\\
  \texttt{evazhu@mit.edu}
  \And
  Rishabh Ranawat\footnotemark[1]\\
  Independent\\
  \texttt{ranawat.rishabh@gmail.com}
}
\hypersetup{pdftitle={What Do RTL Repair Benchmarks Actually Measure?},
  pdfauthor={Yeyin (Eva) Zhu, Rishabh Ranawat}}

\begin{document}
\maketitle

\begin{abstract}
RTL repair benchmarks should measure whether a model can fix broken RTL given a specification. To
this end, we construct and hash-bind RTLRepairDataset, 5{,}784 \emph{(broken, fixed)} training pairs;
RTLRepairBench, 459 held-out cases with 374 syntactic and 85 behavioral mutations; and an evaluation
harness supporting simulation and formal equivalence. We then use this infrastructure to identify two
measurement confounds. First, simulation and formal scoring give opposite
signs for the apparent effect of providing a specification in one evaluated cohort; reset-aware rescoring attributes almost
all of the simulation--formal disagreement to reset and first-cycle comparison semantics. Second,
on 274 mined model failures, the specification-only minus specification-conditioned recovery point
estimate ranges from $-8.5$ to $+18.3$ percentage points across six models. Thus, a conventional repair score
need not isolate whether the model used the faulty RTL. We translate these findings into evaluation
controls: report initialization, reset, first-cycle, and unknown-value semantics; disclose oracle
coverage and undecided cases; and include a specification-only baseline. Together, the resources and
audit provide both a benchmark for behavioral repair and a protocol for diagnosing what contributes
to its measured score. We release the benchmark, the training set, construction code,
sanitized aggregate evidence, and reproduction materials at \url{\releaseurl}.
\end{abstract}

\section{Introduction}

Specification-to-RTL generation has been studied extensively, including work on RTLCoder~\cite{rtlcoder},
OriGen~\cite{origen}, and VeriReason~\cite{verireason}. Repair systems likewise use compiler,
simulation, or testbench feedback to improve a candidate. Behavioral repair remains difficult to
measure, however: a candidate can compile and still disagree with the intended design, and its score
can depend on the oracle contract as well as on which inputs are supplied to the model.

We introduce a tool-gated pipeline and corresponding training and evaluation resources that separate
compiler-diagnosed failures from lint-clean behavioral failures. We then use controlled ablations to
audit two aspects of RTL-repair measurement: whether simulation and formal equivalence assign the
same effect to a prompt intervention, and whether a successful output is a repair of the supplied RTL
or a fresh implementation generated from the specification.

\paragraph{Contributions.}
We make three contributions. First, we construct and release RTLRepairDataset with its tool-gated
builder, and release RTLRepairBench with gates that separate syntactic from behavioral
failures. Second, we
empirically identify two measurement confounds: sensitivity to reset and first-cycle comparison
semantics, and the potential conflation of repair with specification-only regeneration. Third, we
convert these findings into concrete controls for RTL-repair evaluation, including explicit oracle
contracts, oracle-coverage reporting, and specification-only baselines.

\section{Related work}

\paragraph{Generation and syntax repair.}
RTLCoder~\cite{rtlcoder}, OriGen~\cite{origen}, CRAFTRTL~\cite{craftrtl},
CodeV~\cite{codev}, and VerilogDB~\cite{verilogdb} contribute models, data, or training procedures
for spec\to RTL generation. RTLFixer~\cite{rtlfixer} and OriGen's self-reflection use compiler
feedback to repair syntax or compilation failures. AutoChip~\cite{autochip} is broader: it uses both
compiler and simulation feedback, including functional mismatches, to iteratively revise generated
RTL. Our focus is complementary: we construct paired repair data and evaluate lint-clean behavioral
failures under explicitly defined simulation and formal oracles.

\paragraph{Testbench-feedback, debugging, and semantic repair.}
Recent work in VeriReason~\cite{verireason}, CodeV-R1~\cite{codevr1}, and
the method of Wang et al.~\cite{verilogprefer} improves functional \emph{generation} using
testbench-derived rewards or preferences. VeriDebug~\cite{veridebug} combines contrastive
representations with guided Verilog correction, ARSP~\cite{arsp} targets functional repair through
semantic partitioning, and VeriRAG~\cite{verirag} addresses the distinct problem of RTL repair for
design-for-test compliance. RTLRepairBench differs by coupling a released paired repair resource
with an audit of how oracle and input-condition choices change the measured outcome.

\section{RTLRepairDataset, RTLRepairBench, and evaluation design}\label{sec:dataset}

\begin{figure}[H]
\centering
\includegraphics[width=0.94\linewidth]{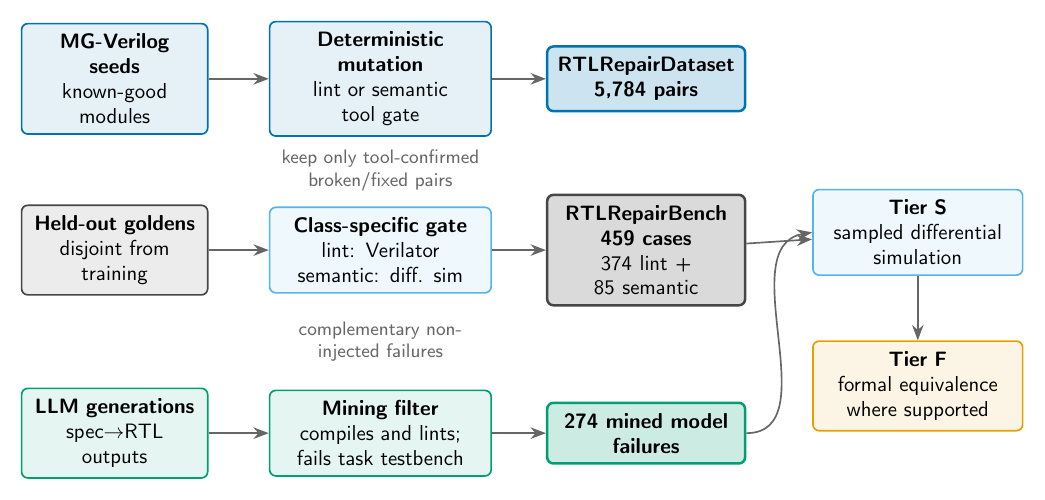}
\caption{Tool-gated construction and evaluation. Training and held-out resources use disjoint seeds;
mined model failures provide a complementary check beyond injected mutations.}
\label{fig:pipeline}
\end{figure}

Figure~\ref{fig:pipeline} summarizes the pipeline, and Appendix~\ref{app:mutations} reports the exact
split of the injected mutants. We ensure that RTLRepairDataset and RTLRepairBench use
disjoint source seeds and that each broken/fixed pair is kept only when its class-specific gate
confirms the injected failure. RTLRepairDataset contains 5{,}784 training pairs; RTLRepairBench
contains 459 held-out cases, including four semantic operators unseen in training.

Training seeds originate from MG-Verilog~\cite{mgverilogpaper,mgverilog}; the held-out and decontamination pool uses
VerilogEval~\cite{verilogeval} and RTLLM~\cite{rtllm}. Normalized seed identifiers show
no overlap between 1{,}895 distinct historical training seeds and the benchmark's 130 distinct seeds.
Appendix~\ref{app:provenance} reports the surviving snapshot and licensing records and their limits.

\paragraph{Evaluation harness and oracle definitions.}
We have built an open-source evaluation harness that supports two types of oracles:
\textbf{Tier S: Simulation} and \textbf{Tier F: Formal Equivalence}. Tier S runs the candidate and
golden module for 200 cycles under identical sampled stimuli and checks their outputs at every
compared cycle. The harness uses \code{verilator}~\cite{verilator} for linting and
\code{iverilog}~\cite{iverilog} for simulation. An observed mismatch is a counterexample under that
trace and simulation contract, whereas agreement over 200 sampled cycles does not prove equivalence.
Tier F applies the contract matched to each design: exhaustive combinational checking (17 cases), a
declared reset sequence (60), or a designated preamble (8), followed by equivalence checking under
stated initialization and unknown-value semantics. These assumptions matter when state is not fully
initialized~\cite{eqyxprop}.

In the six-model subset selected for syntactic-repair recovery between $87.7\%$ and $100.0\%$,
behavioral recovery ranged descriptively from $10.6\%$ to $75.3\%$. This subset exhibits a substantial gap
while retaining the bug classes and oracle outcomes needed to investigate its source.

\paragraph{Mined failures.}\label{mined}
Synthetic mutations could introduce construction-specific bias. We therefore mine \emph{model-generated}
failures for which the spec\to RTL output compiles but fails the task testbench. We use this cohort to
probe the relationship between specification and regeneration for repair tasks in
Section~\ref{section5}.

\section{Oracle sensitivity: the estimated specification effect reverses across contracts}

To isolate the effect of the tool diagnostic and the specification provided to the models, we ran
the evaluation with both oracles. This evaluation was restricted to the semantic candidates with
definitive formal-equivalence verdicts (79 out of 85). We present the results for
GPT-5.5 here for brevity, although we ran experiments across a variety of models.

\begin{table}[H]
\centering
\scriptsize
\setlength{\tabcolsep}{3.5pt}
\caption{Same 79 cases (one retained candidate per arm), restricted to cases with definitive Tier~F
verdicts in both arms. Entries are passes/79 (percent); contrasts are $+$specification minus diagnostic
from unrounded counts. Cycle~1 skips only the first comparison; reset-aware simulation applies the
released two-edge reset heuristic and remains a bounded sampled test.}
\label{tab:oracle}
\begin{tabular}{lrrr}
\toprule
Scoring contract & Diagnostic & $+$ specification & Paired contrast \\
\midrule
Tier~S, cycles 0--199 & 60/79 (75.9\%) & 51/79 (64.6\%) & $-11.4$ pp \\
Tier~S, cycles 1--199 & 72/79 (91.1\%) & 78/79 (98.7\%) & $+7.6$ pp \\
Tier~S, reset-aware initialization & 71/79 (89.9\%) & 78/79 (98.7\%) & $+8.9$ pp \\
Tier~F, per-design equivalence & 71/79 (89.9\%) & 78/79 (98.7\%) & $+8.9$ pp \\
\bottomrule
\end{tabular}
\end{table}

Skipping only the first comparison reverses the descriptive contrast from $-11.4$ to $+7.6$ points.
Reset-aware initialization changes one additional diagnostic-arm verdict; on this 79-case cohort its
158 verdicts coincide with Tier~F. This agreement supports sensitivity to initialization and
first-cycle semantics, but it does not make bounded simulation equivalent to formal proof.

This is an exploratory complete-case comparison from one greedy draw per arm, not a population-level
effect estimate. Six of 85 cases lack a definitive Tier~F verdict and are excluded; candidates also
share source designs, so the 79 case outcomes should not be treated as independent replicates.

Within the definitive 79-case cohort, reset-aware rescoring cleared all 39 Tier-S-fail/Tier-F-proved
outcomes (12 diagnostic-arm and 27 specification-arm). Across all 85 cases, it cleared 12/12 and
27/30 in the two arms, respectively. Reset and first-cycle semantics
therefore explain almost all, but not every, observed simulation--formal disagreement. For benchmarks
that rely on simulation matches~\cite{verilogeval,rtllm,autochip},
the reset sequence and first compared cycle are therefore part of the measurement definition and
should be reported and audited.

\section{Separating repair from specification-only regeneration}\label{section5}

To test whether output correctness alone separates repair from fresh generation, each model received
one of three inputs: (1) broken RTL without the specification, (2) broken RTL with the specification,
or (3) the specification alone. The third arm is a necessary but not sufficient diagnostic: aggregate
similarity between (2) and (3) cannot reveal whether an individual output used the broken RTL. We ran
this evaluation on the 274-case mined dataset described in Section~\ref{mined}. Exact prompt templates
and the surviving call and scoring settings are included in the released artifact.

Table~\ref{tab:regen} reports the six-model exploratory ablation after guarding the
pre-initialization comparison. Each model starts from the same 274 failures; $n$ is the intersection
with a retained, scoreable candidate in all three arms. We report point estimates rather than nominal
case-level $p$ values because cases cluster within 112 source tasks and the panel was not designed for
multiplicity-adjusted confirmatory inference.

\begin{table}[H]
\centering
\scriptsize
\setlength{\tabcolsep}{3.5pt}
\renewcommand{\arraystretch}{1.05}
\caption{Cycle-1 complete-case regeneration control. $\Delta_{\mathrm{c1}}$ is specification-only
minus specification-conditioned repair. Model names identify historical endpoint-output snapshots,
not immutable provider revisions.}
\label{tab:regen}
\begin{tabular}{@{}lrrrrr@{}}
\toprule
Model (family) & $n$ & No spec & $+$spec & Spec only & $\Delta_{\mathrm{c1}}$ \\
\midrule
GPT-5.5 (OpenAI) & 260 & 57.3\% & 81.5\% & 76.9\% & $-4.6$ pp \\
Claude Opus 4.8 (Anthropic) & 271 & 34.3\% & 71.6\% & 69.0\% & $-2.6$ pp \\
GPT-5.4-mini (OpenAI) & 274 & 29.6\% & 51.5\% & 56.9\% & $+5.5$ pp \\
GPT-4.1 (OpenAI) & 271 & 27.3\% & 46.5\% & 38.0\% & $-8.5$ pp \\
GPT-4o-mini (OpenAI) & 274 & 19.0\% & 39.1\% & 41.2\% & $+2.2$ pp \\
Claude Opus 4.6 (Anthropic) & 273 & 9.5\% & 51.6\% & 70.0\% & $+18.3$ pp \\
\bottomrule
\end{tabular}
\end{table}

Specification-only recovery exceeds no-specification recovery for every model. Relative to
specification-conditioned repair, however, the observed specification-only point estimate is lower
for three models and higher for three, with $\Delta_{\mathrm{c1}}$ ranging from $-8.5$ to $+18.3$
points. These results do not establish that a model ignored the faulty RTL internally; they show that
output correctness alone cannot identify whether success arose from repair or regeneration. The
specification-only arm exposes this ambiguity but does not resolve the mechanism.

\section{Conclusion}

Tool-diagnosed syntactic errors are easier for models to fix than behavioral bugs that compile
cleanly for the evaluated models. RTLRepairDataset and RTLRepairBench make it possible to quantify this
gap through a released, hash-bound training resource and its builder, a held-out evaluation
set, mutation labels, and explicit oracle outcomes. The audit
also shows that simulation and formal-equivalence scores can disagree substantially, with reset and
first-cycle semantics explaining almost all observed false divergences, and that a correct output need
not demonstrate use of the supplied faulty RTL. More explicit oracle contracts and regeneration
controls can therefore reduce misleading RTL-repair claims and improve reproducibility.

These findings motivate four reporting controls for RTL repair benchmarks: (1) separate syntactic
from behavioral bugs; (2) state initialization, reset, first-comparison-cycle, and unknown-value
semantics; (3) include a specification-only baseline as a diagnostic for possible regeneration;
and (4) report oracle coverage and undecided cases. The baseline in (3) is necessary but not sufficient
for mechanism attribution. These controls make the interpretation of a
repair score reproducible alongside the score itself. A repair system or benchmark score can
nevertheless create false confidence if deployed as a substitute for verification; generated RTL
should continue to undergo independent simulation, formal analysis where supported, and human
design review.

\paragraph{Limitations and Future Work.}
Our findings are descriptive and limited by the evaluated cohort sizes, a predominantly single-draw
setting, and historical model-output snapshots. Future work should test repeated samples, a broader
model roster, and larger and more diverse RTL designs to assess how well the findings generalize.

\raggedbottom

\appendix
\section{Oracle definitions}\label{app:oracles}

\begin{figure}[H]
\centering
\includegraphics[width=\linewidth]{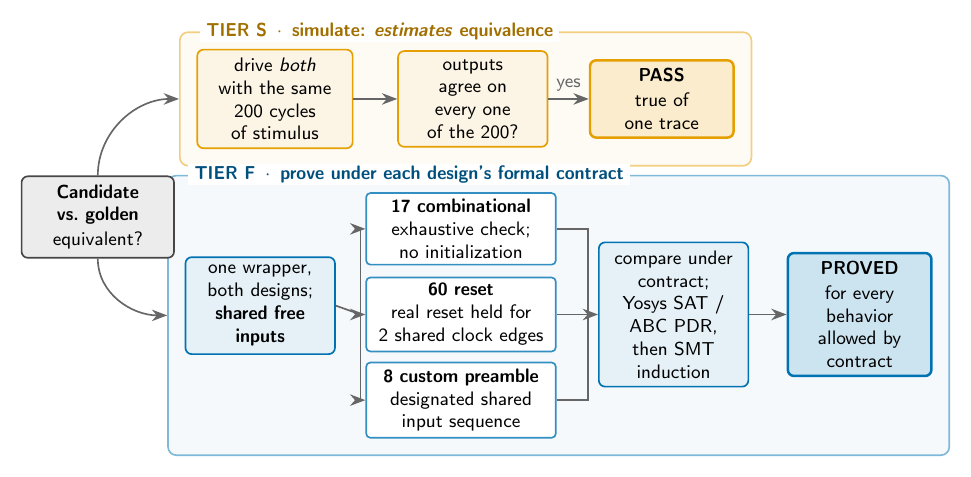}
\caption{High-level structure of the Tier~S and Tier~F oracles. Tier~S agreement is evidence for one
sampled trace; Tier~F applies a design-matched combinational, reset, or preamble contract before
attempting a proof with Yosys~\cite{yosys} SAT, ABC~\cite{abc} PDR, and SMT induction.}
\label{fig:oracles}
\end{figure}

\section{Mutation distribution}\label{app:mutations}

\begin{table}[H]
\centering
\footnotesize
\setlength{\tabcolsep}{4pt}
\renewcommand{\arraystretch}{1.05}
\caption{Mutation inventory and validation gates. A dash marks an operator absent from that split.
RTLRepairBench includes four semantic operators that do not appear in RTLRepairDataset.}
\label{tab:mutations}
\begin{tabular}{@{}l>{\raggedright\arraybackslash}p{0.52\textwidth}rrl@{}}
\toprule
bucket & operator (what it breaks) & train & bench & validation gate \\
\midrule
\multirow{6}{*}{lint}
 & \code{drop\_endmodule} --- deletes the closing \code{endmodule}
   & 1{,}000 & 130 & \multirow{6}{*}{\code{verilator -{}-lint-only}} \\
 & \code{delete\_semicolon} --- drops a statement terminator
   & 1{,}000 & 130 & \\
 & \code{drop\_one\_end} --- unbalances a \code{begin}/\code{end} block
   & 1{,}000 & 55 & \\
 & \code{rename\_declaration} --- renames a declaration; its uses go undeclared
   & 1{,}000 & 23 & \\
 & \code{blocking\_in\_seq} --- \code{=} for \code{<=} in a clocked block
   & 1{,}000 & 36 & \\
 & \code{width\_mismatch} --- pins a parameterized vector to \code{[3:0]}
   & 403 & --- & \\
\midrule
\multirow{6}{*}{semantic}
 & \code{op\_swap} --- \code{+}\,$\leftrightarrow$\,\code{-} and \code{\&}\,$\leftrightarrow$\,\code{|}
   & 300 & 20 & \multirow{6}{*}{differential \code{iverilog} sim} \\
 & \code{flip\_reset\_polarity} --- drops the \code{!} from an active-low reset test
   & 81 & 2 & \\
 & \code{flip\_equality} --- \code{==}\,$\leftrightarrow$\,\code{!=}
   & --- & 22 & \\
 & \code{swap\_ternary} --- swaps the two arms of a mux
   & --- & 20 & \\
 & \code{swap\_logical} --- \code{\&\&}\,$\leftrightarrow$\,\code{||}
   & --- & 17 & \\
 & \code{off\_by\_one\_literal} --- nudges a sized literal by $\pm 1$
   & --- & 4 & \\
\midrule
total & & 5{,}784 & 459 & \\
\bottomrule
\end{tabular}
\end{table}

\section{Data provenance and historical model metadata}\label{app:provenance}

\begin{table}[H]
\centering
\footnotesize
\setlength{\tabcolsep}{4pt}
\caption{Surviving upstream records. ``MIT'' is the upstream repository or dataset-card declaration,
not a project grant over every derived row.}
\begin{tabular}{@{}llll@{}}
\toprule
Asset & surviving revision record & role & release boundary \\
\midrule
MG-Verilog & \code{82944845640b...} (dataset snapshot), MIT & training seeds & training file released \\
VerilogEval & \code{c498220} (abbreviated), MIT & held-out/mined tasks & benchmark released \\
RTLLM & \code{41b2689} (abbreviated), MIT & decontamination/legacy probes & no legacy probe claim \\
\bottomrule
\end{tabular}
\end{table}

The historical training output contains 1{,}895 distinct seeds; the released benchmark contains 130.
We normalized seed identifiers before splitting and found zero overlap. The released training file
has 5{,}775 pairs: relative to Table~\ref{tab:mutations}, it lacks 6 \code{rename\_declaration} and 3
\code{width\_mismatch} pairs. Table~\ref{tab:oracle}'s archive records the served alias
\code{azure/openai/gpt-5.5}, greedy decoding, and a matched 4{,}096-token output budget. For the
cross-model panel, retained scored outputs and aggregates survive, but exact immutable provider
revisions, call times, and some request metadata do not. Consequently, all model labels denote
historical endpoint-output snapshots rather than reproducible model versions.

\end{document}